\documentclass[conference]{IEEEtran}
\IEEEoverridecommandlockouts
\usepackage{amsmath,amssymb,amsfonts}
\usepackage{algorithmic}
\usepackage{graphicx}
\usepackage{textcomp}
\usepackage{subcaption}
\usepackage{hyperref}

\usepackage{booktabs}
\usepackage{multirow}
\usepackage{xcolor}
\usepackage[backend=biber, style=ieee,doi=false,isbn=false,url=false]{biblatex}
\def\BibTeX{{\rm B\kern-.05em{\sc i\kern-.025em b}\kern-.08em
    T\kern-.1667em\lower.7ex\hbox{E}\kern-.125emX}}
\begin{document}

\title{DO-RAG: A Domain-Specific QA Framework Using Knowledge Graph-Enhanced Retrieval-Augmented Generation\\
}

\author{
	\IEEEauthorblockN{
		David Osei Opoku\IEEEauthorrefmark{1}, 
		Ming Sheng\IEEEauthorrefmark{2}, 
		Yong Zhang\thanks{* Yong Zhang (zhangyong05@tsinghua.edu.cn) is the corresponding author.
		}\IEEEauthorrefmark{3}
		} 
	\IEEEauthorblockA{\IEEEauthorrefmark{1}Department of Computer Science and Technology, Tsinghua University, Beijing, China}
	\IEEEauthorblockA{\IEEEauthorrefmark{2,3}Beijing National Research Center for Information Science and Technology - Tsinghua University, Beijing, China}

} 

\maketitle

\begin{abstract}
Domain-specific QA systems require not just generative fluency but high factual accuracy grounded in structured expert knowledge. While recent Retrieval-Augmented Generation (RAG) frameworks improve context recall, they struggle with integrating heterogeneous data and maintaining reasoning consistency. To address these challenges, we propose \textbf{DO-RAG}, a scalable and customizable hybrid QA framework that integrates multi-level knowledge graph construction with semantic vector retrieval. Our system employs a novel agentic chain-of-thought architecture to extract structured relationships from unstructured, multimodal documents, constructing dynamic knowledge graphs that enhance retrieval precision. At query time, DO-RAG fuses graph and vector retrieval results to generate context-aware responses, followed by hallucination mitigation via grounded refinement. Experimental evaluations in the database and electrical domains show near-perfect recall and over 94\% answer relevancy, with DO-RAG outperforming baseline frameworks by up to 33.38\%. By combining traceability, adaptability, and performance efficiency, DO-RAG offers a reliable foundation for multi-domain, high-precision QA at scale.
\end{abstract}


\section{Introduction}

Question-answering (QA) systems enable users to retrieve information from large corpora by posing natural-language queries. They fall into two broad categories: open-domain QA, which draws on general knowledge to answer questions, and closed-domain QA, which relies on specialized, often technical, resources to address queries. Closed-domain scenarios expose the limitations of generic models, which must be augmented with structured expert knowledge to deliver precise, reliable answers.

Recent advances in large language models (LLMs) such as DeepSeek-R1~\cite{deepseekai2025}, Grok 3~\cite{xai2025grok3}, QwQ-32B~\cite{qwen-qwq}, and OpenAI's O3~\cite{openai2024o3} have dramatically improved fluency and contextual understanding. Yet these models depend predominantly on parametric knowledge and can hallucinate or provide imprecise responses when confronted with domain-specific terminology or complex multi-step reasoning~\cite{yang-etal-2025-curiousllm, zhou2024llmtest}.

To address these limitations, Retrieval-Augmented Generation (RAG) has emerged as a popular approach for improving factual consistency by retrieving relevant text snippets before generation~\cite{siriwardhana-etal-2023-improving, 9072442, bhushan-etal-2025-systematic}. In parallel, knowledge graphs (KGs) offer a complementary solution by encoding entities and their relations in a structured form, facilitating multi-hop reasoning and precise context assembly~\cite{ZhangAndWu2023, tu-etal-2025-lightprof}.

However, existing RAG frameworks struggle to capture the intricate relationships among entities in technical manuals and multimodal resources, leading to fragmented retrieval outputs and persistent hallucinations~\cite{Wang2024KnowledgeGraphPrompting, zafar2024kimedqa, dagnino2024dataquality}. When retrieval and generation components remain loosely coupled, there is no guarantee that the final answer faithfully reflects the retrieved evidence. Although knowledge graphs promise more structured context, manually constructing and maintaining high-quality, domain-specific graphs is labor-intensive, and marrying them with vector search and LLM prompting adds substantial engineering overhead. Consequently, current KG-RAG hybrids face scalability bottlenecks and require extensive manual tuning to remain robust as knowledge evolves.

In this paper, we introduce \textbf{DO-RAG}, a \textbf{Do}main-specific \textbf{R}etrieval-\textbf{A}ugmented \textbf{G}eneration framework that end-to-end automates the transformation of unstructured, multimodal documents into a dynamic, multi-level knowledge graph via an agentic chain-of-thought extraction pipeline; tightly fuses the resulting graph traversal with semantic vector search to assemble a context-rich prompt; and then grounds generation in retrieved evidence through a refinement step that detects and corrects hallucinations.

Our contributions are as follows:
\begin{itemize}
  \item \textbf{Agentic, multi-stage KG construction.} We design and implement a hierarchical, agent-based extraction pipeline that processes text, tables, code snippets, and images to automatically build and update a knowledge graph capturing entities, relations, and attributes.
  \item \textbf{Hybrid retrieval fusion.} We develop a unified mechanism that merges graph-based traversal with semantic search at query time, ensuring that all relevant, structurally grounded information informs the LLM’s prompt.
  \item \textbf{Grounded hallucination mitigation.} We introduce a post-generation refinement step that cross-verifies initial LLM outputs against the knowledge graph and iteratively corrects inconsistencies, dramatically reducing factual errors.
  \item \textbf{Plug-and-play modularity.} Our framework accommodates diverse LLMs and retrieval modules, enabling seamless component swapping and straightforward extension to new domains without retraining.
\end{itemize}

\noindent Evaluated on expert-curated benchmarks in the database and electrical engineering domains, DO-RAG achieves near-perfect contextual recall and over 94\% answer relevancy, outperforming existing RAG platforms such as FastGPT~\cite{fastgpt2025}, TiDB.AI~\cite{tidbai} and Dify.AI~\cite{difyai} by up to 33.38\%. By unifying structured and generative approaches, DO-RAG provides a scalable, high-precision solution for domain-specific QA.

The rest of this paper is organized as follows: Section~\ref{sec2:related_work} reviews existing work on domain-specific QA and RAG frameworks. Section~\ref{sec3:approach} explains the design and approach of DO-RAG. Section~\ref{sec4:experiment} presents the experimental setup, datasets, metrics, and results, comparing DO-RAG to baselines, and ends with a discussion of limitations and future work. Section~\ref{sec5:conclusion} summarizes the main findings and contributions.

\section{Preliminary and Related Work}\label{sec2:related_work}

\subsection{Preliminary}
In domain-specific question-answering, a system retrieves answers from a document corpus \(\mathcal{D}\) and a knowledge graph \(\mathcal{G} = (V, E, W)\), where \(V\) is a set of entities, \(E\) denotes relationships, and \(W: E \to [0,1]\) assigns confidence weights. A query \(q\) is mapped to a vector:
\begin{equation}
Q = E(q), \quad Q \in \mathbb{R}^d,
\end{equation}
\noindent where \(E\) is an embedding function, and \(d\) is the embedding dimension. Document chunks \(c_i \in \mathcal{D}\) are similarly embedded:
\begin{equation}
C_i = E(c_i), \quad C_i \in \mathbb{R}^d.
\end{equation}
\noindent The retrieval score \(S\) combines semantic and graph-based relevance:
\begin{equation}
S = \alpha \cdot \max_{i} \text{sim}(Q, C_i) + (1 - \alpha) \cdot R(\mathcal{G}_Q),
\end{equation}
where \(\text{sim}(Q, C_i)\) is cosine similarity, \(R(\mathcal{G}_Q)\) retrieves relevant entities from \(\mathcal{G}\), and \(\alpha \in [0,1]\) balances contributions. The answer is generated as:
\begin{equation}
A = G(Q, S),
\end{equation}
\noindent where \(G\) is the generation function. The goal is to minimize factual errors:
\begin{equation}
\min_{\theta} \mathbb{E}_{(q, a^*) \sim \mathcal{D}_{\text{eval}}} \mathcal{L}(G(Q, S; \theta), a^*),
\end{equation}
\noindent with \(\theta\) as model parameters, \(\mathcal{D}_{\text{eval}}\) as the evaluation dataset with ground-truth answers \(a^*\), and \(\mathcal{L}\) as the loss function.

\subsection{Related Work}
Domain-specific QA systems have evolved from rule-based approaches to advanced methods integrating structured knowledge and generative models. Early systems relied on static rules and limited knowledge bases, restricting their adaptability to complex queries \cite{allam2012question}. Knowledge graphs address this by encoding entities and relationships, enhancing reasoning, but their manual construction is resource-intensive \cite{dimitrakis2020survey}.

Large language models excel in natural language tasks but struggle in specialized domains due to hallucinations and insufficient domain knowledge \cite{singhal2023large}. Retrieval-augmented Generation mitigates this by grounding responses in retrieved documents \cite{lewis2020retrieval}, yet faces challenges in capturing intricate relationships in technical contexts \cite{siriwardhana-etal-2023-improving, zhu2025knowledge}.

Current frameworks like KIMedQA \cite{zafar2024kimedqa} and Panda \cite{singh2024panda} are domain-specific but lack scalability and flexibility. KIMedQA achieves high medical QA accuracy but is computationally intensive, while Panda, focused on database debugging, does not support multi-domain applications. KG-RAG \cite{soman2023biomedical} integrates biomedical KGs with RAG but struggles with scalability.

To overcome these limitations, we propose DO-RAG, a framework that automates dynamic KG construction, fuses vector retrieval with grounded generation, and enhances precision and adaptability in domain-specific QA.

\section{Approach}\label{sec3:approach}
\subsection{System Overview}

\begin{figure}[ht!]
  \centering
  \includegraphics[width=1.0\linewidth]{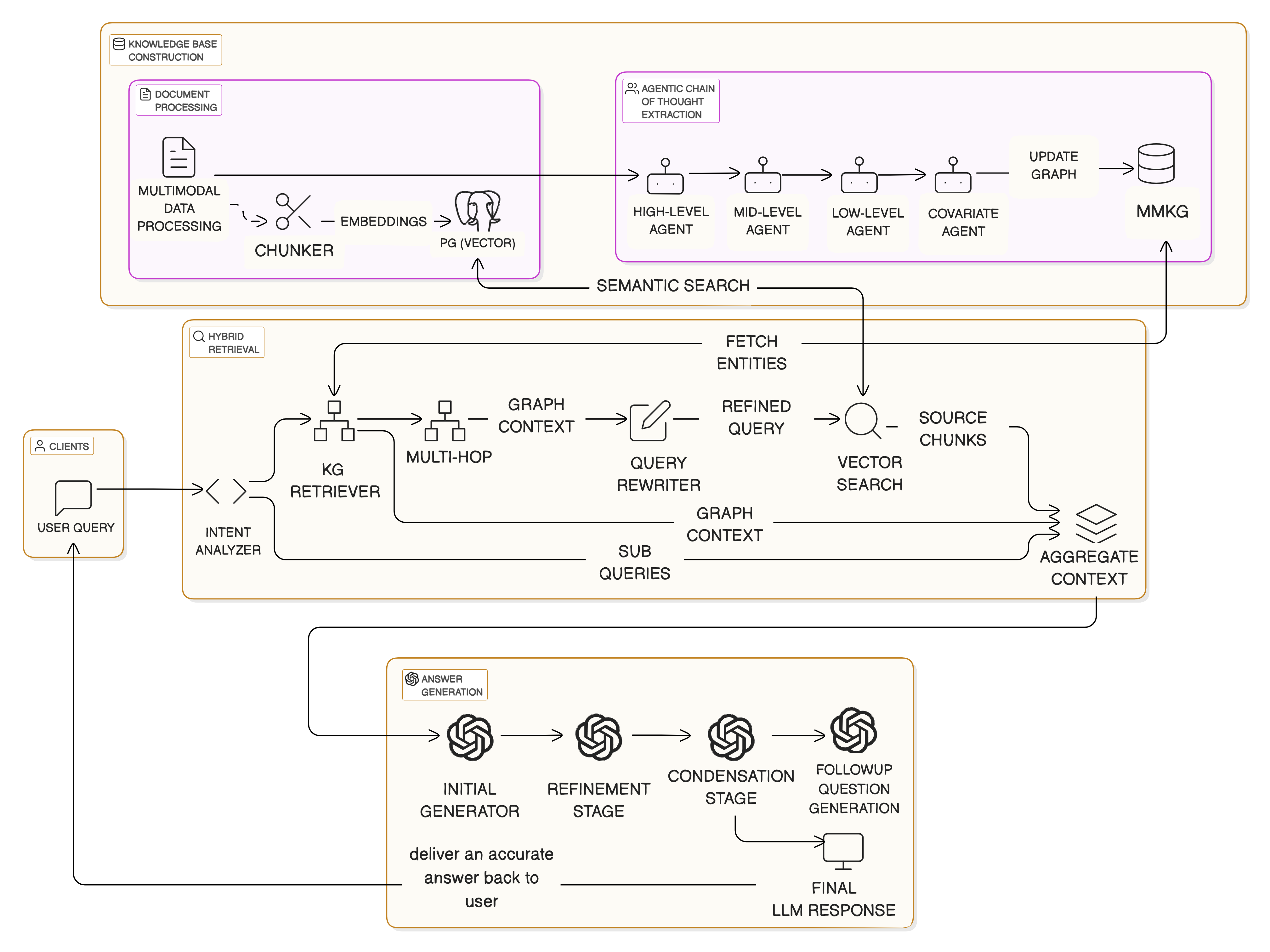}
  \caption{High-level overview of DO-RAG methodology.}
  \label{fig:overview_pipeline}
\end{figure}

As illustrated in Figure~\ref{fig:overview_pipeline}, DO-RAG comprises four key stages: (1) multimodal document ingestion and chunking, (2) multi-level entity-relation extraction for knowledge graph (KG) construction, (3) hybrid retrieval combining graph traversal and dense vector search, and (4) a multi-stage generation pipeline for grounded, user-aligned answers.

It begins by parsing heterogeneous domain data---such as logs, technical manuals, diagrams, and specifications---into meaningful chunk units. These chunks are stored alongside their vector embeddings in a pgvector-enabled PostgreSQL instance. Simultaneously, an agentic chain-of-thought entity extraction pipeline transforms document content into a structured multimodal knowledge graph (MMKG), capturing multi-granular relationships such as system parameters, behaviors, and dependencies.

At the point of user interaction, the user query undergoes decomposition through an intent recognition module, which segments it into one or more sub-queries representing discrete informational intents. The system first performs retrieval from the knowledge graph by embedding the initial query and matching it to relevant entities. This is followed by a multi-hop traversal to expand the retrieval scope, yielding structured, domain-specific context grounded in entity relationships and metadata.

Next, this graph-derived context is used to rewrite the original query into a more specific and disambiguated form using a graph-aware prompt template. The refined query is then encoded into a dense vector and used to retrieve top-k semantically similar chunks from the vector database.

At this point, the system consolidates all relevant information sources: the original user query, its refined version, the knowledge graph context, the retrieved text chunks, and prior user interaction history. These components are integrated into a unified prompt structure and passed into the generation pipeline.

The generation proceeds in three stages: initial answer generation, refinement for factual consistency and clarity, and condensation to ensure coherence and brevity. Finally, DO-RAG invokes a follow-up question generator, which formulates next-step queries based on the refined answer and overall conversation context, enhancing user engagement and supporting multi-turn interaction.
\subsection{Knowledge Base Construction}
Document processing begins with multimodal ingestion, where text, tables, and images are normalized and segmented into coherent chunks. Metadata---including source file structure, section hierarchy, and layout tags—is retained for traceability.

\begin{figure}[ht!]
  \centering
  \includegraphics[width=1.0\linewidth]{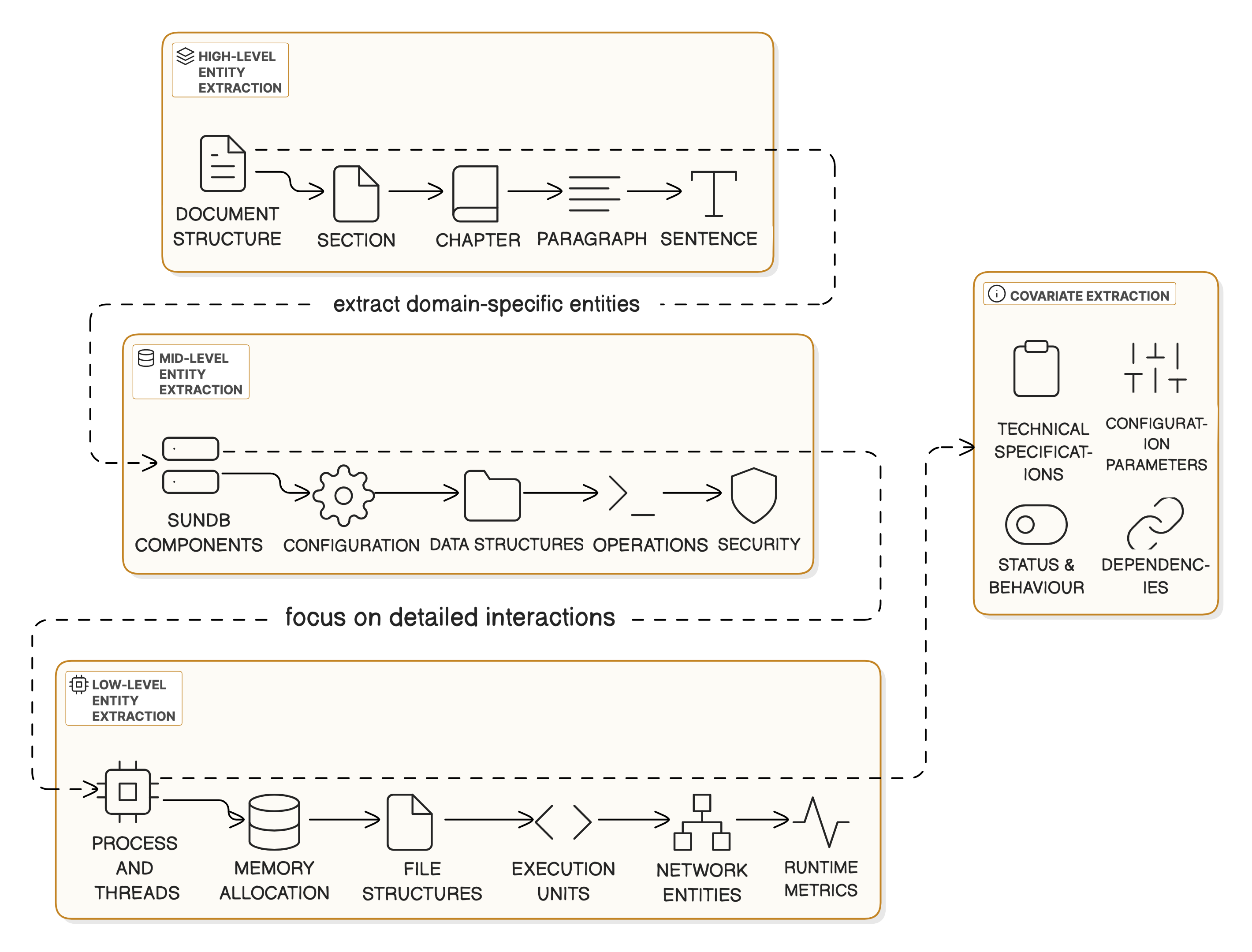}
  \caption{Multi-level entity-relation extraction.}
  \label{fig:extraction_agents}
\end{figure}

In parallel, we extract structured knowledge using a multi-agent, multi-level pipeline. As shown in Figure~\ref{fig:extraction_agents}, the pipeline includes four specialized agents operating at different abstraction levels:
\begin{itemize}
  \item \textbf{High-Level Agent}: Identifies structural elements (e.g., chapters, sections, paragraphs).
  \item \textbf{Mid-Level Agent}: Extracts domain-specific entities such as system components, APIs, and parameters.
  \item \textbf{Low-Level Agent}: Captures fine-grained operational relationships like thread behavior or error propagation.
  \item \textbf{Covariate Agent}: Attaches attributes (e.g., default values, performance impact) to existing nodes.
\end{itemize}

The output is a dynamic KG \( \mathcal{G} = (V, E, W) \), where nodes \( V \) represent entities, edges \( E \) represent relationships, and weights \( W \) encode confidence scores. To avoid redundancy, deduplication is enforced using cosine similarity between new and existing entity embeddings. Additionally, synopsis nodes are synthesized to group similar entities and reduce graph complexity.

\subsection{Hybrid Retrieval and Query Decomposition}

\begin{figure}[ht!]
    \centering

    \begin{subfigure}{1\linewidth}
        \centering
        \includegraphics[height=0.50\linewidth]{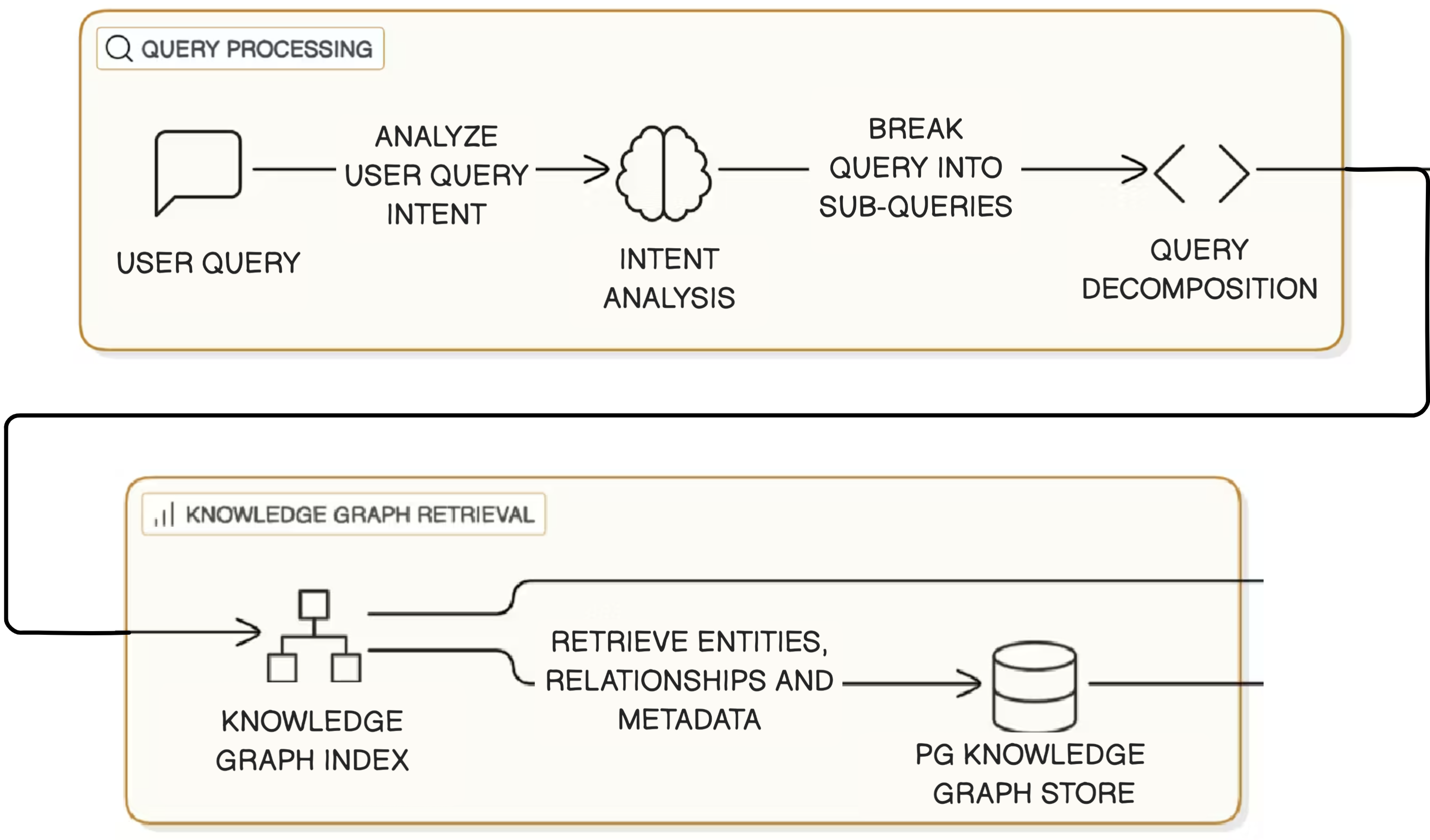}
        \caption{Query processing and knowledge graph retrieval.}
        \label{fig:query_processing_kg_retrieval}
    \end{subfigure}
    
    \vspace{0.5em} 

    \begin{subfigure}{1\linewidth}
        \centering
        \includegraphics[height=0.73\linewidth]{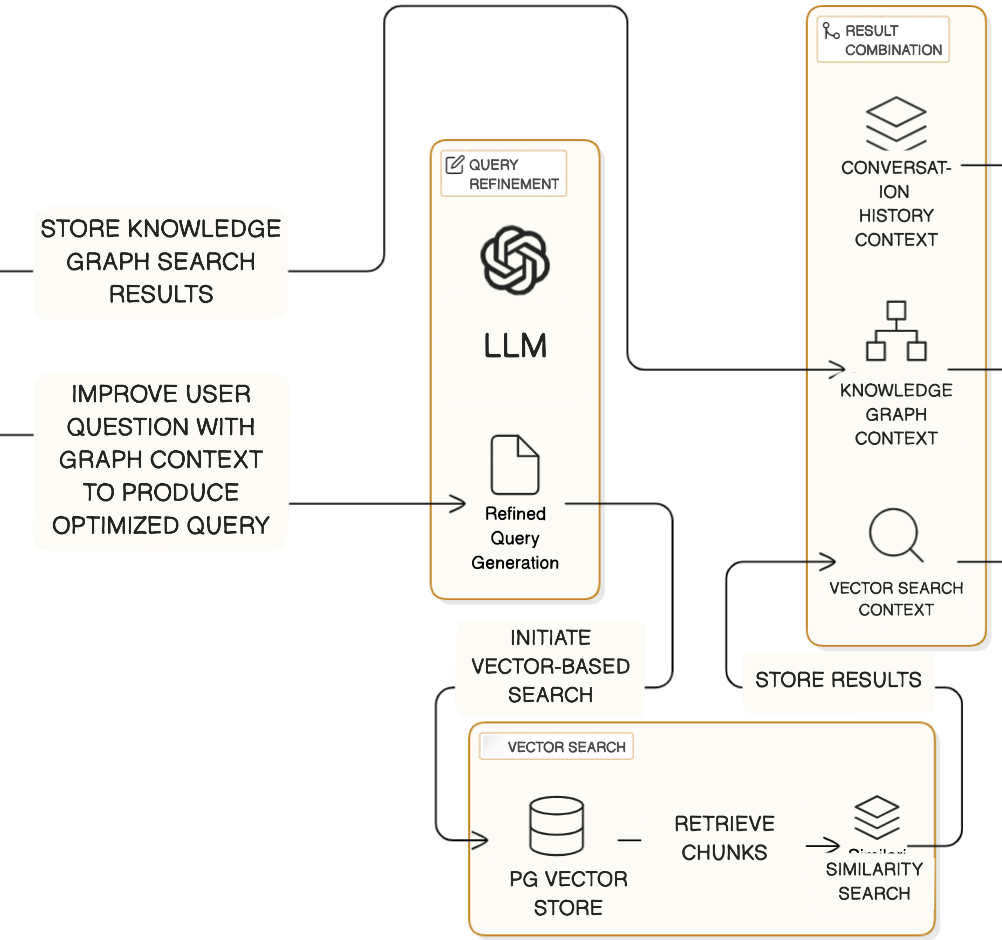}
        \caption{Query refinement, vector Search, and result combination.}
        \label{fig:query_refinement_vector_search}
    \end{subfigure}
    \caption{Overview of the retrieval process.}
    \label{fig:retrieval_overview}
\end{figure}

 As illustrated in Figure~\ref{fig:retrieval_overview},when a user submits a question, DO-RAG performs a structured decomposition using an LLM-based intent analyzer. This yields sub-queries that are used to guide retrieval from both the KG and the vector store. It first retrieves relevant nodes from the KG using semantic similarity, then performs multi-hop traversal to assemble a context-rich subgraph. This graph-derived evidence is used to rewrite and disambiguate the query via a graph-aware prompt. The refined query is then vectorized and used to retrieve semantically similar chunks from the vector database. Finally, DO-RAG aggregates all sources---original and refined queries, graph context, vector-retrieved chunks, and user conversation history---into a unified prompt structure.

\subsection{Grounded Answer Generation and Delivery}

\begin{figure}[ht!]
    \centering

    \begin{subfigure}{0.99\linewidth}
        \centering
        \includegraphics[width=\linewidth]{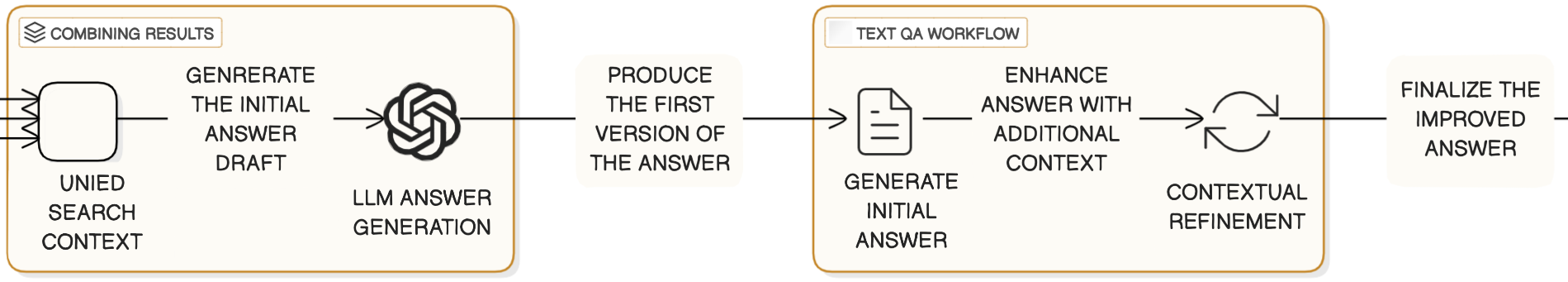}
        \caption{Combining results and Text QA workflow.}
        \label{fig:combining_results_text_qa_workflow}
    \end{subfigure}

    \vspace{1em} 

    \begin{subfigure}{0.99\linewidth}
        \centering
        \includegraphics[width=\linewidth]{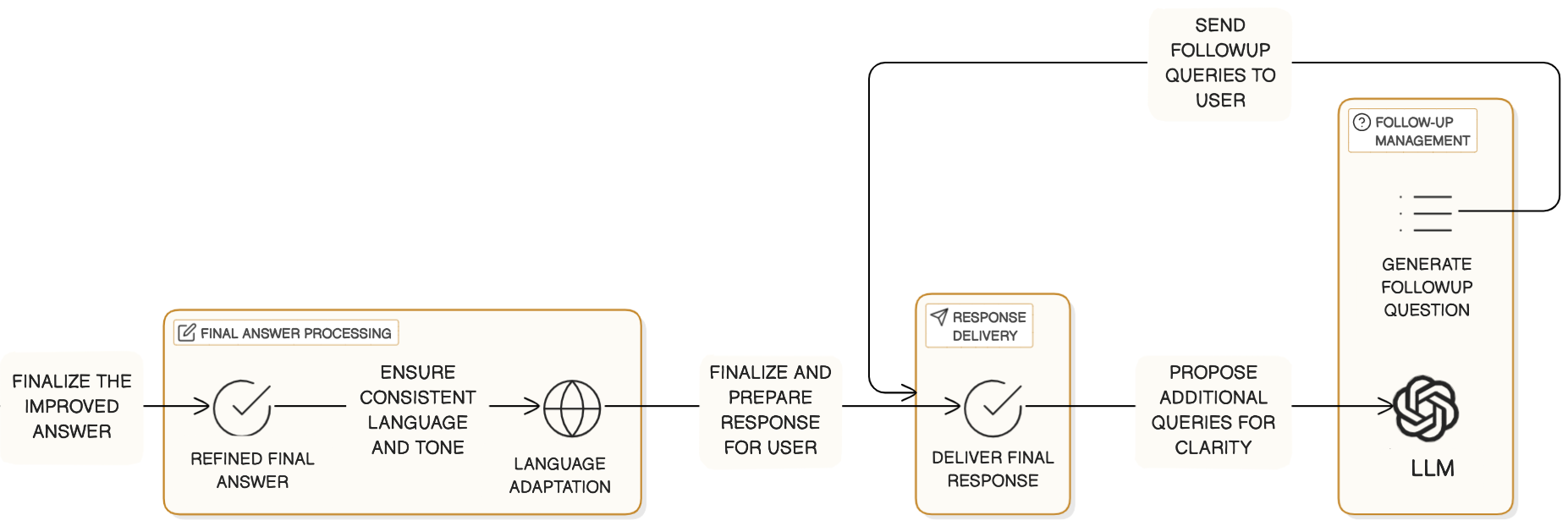}
        \caption{Final answer processing, response delivery, and follow-up management.}
        \label{fig:final_answer_processing_response_delivery_followup}
    \end{subfigure}

    \caption{Overall response pipeline.}
    \label{fig:response_delivery}
\end{figure}

As shown in Figure~\ref{fig:response_delivery}, the final answer is generated using a staged prompt strategy. The initial naive prompt instructs the LLM to answer based only on the retrieved evidence while explicitly avoiding unsupported content. The output is passed through a refinement prompt that restructures and validates the answer, followed by a condensation stage that aligns tone, language, and style with the original query.

To enhance user engagement and simulate expert-like guidance, DO-RAG also generates follow-up questions based on the refined answer. The final output includes (1) a polished, verifiable answer, (2) citation footnotes tracing the answer to the source, and (3) a set of targeted follow-up questions. If the system cannot locate sufficient evidence, the model is required to return “I do not know,” preserving reliability and preventing hallucination.

\section{Experiments}\label{sec4:experiment}

To evaluate the DO-RAG framework, we selected SunDB, a distributed relational database management system developed by Client Service International, Inc. (CSII), as the specialized domain. SunDB is widely used in critical industries such as banking, telecommunications, and government, offering a complex, heterogeneous dataset of technical manuals, system logs, and specifications. This domain is ideal for testing DO-RAG’s multimodal ingestion, multi-level entity-relationship extraction, hybrid retrieval, and grounded answer generation.

\subsection{Experimental Setup}

\textbf{Hardware and Software. }The evaluation was conducted on a high-performance Ubuntu workstation equipped with 64GB RAM, an NVIDIA A100 80GB GPU, and a 1TB hard disk. The experiment was implemented using locally deployed software to prioritize data security and efficiency. LangFuse (v3.29.0)~\cite{langfuse} enabled tracing and logging of retrieval-augmented generation operations, Redis (v7.2.5)~\cite{redis2025} managed caching, MinIO (v2025-04-22T22-12-26Z)~\cite{minio2025} handled document storage, and ClickHouse (v25.4.3.22-stable)~\cite{clickhouse2025} supported high-performance analytics. PostgreSQL (v16.4)~\cite{postgresql2025} served as the relational backend storage with pgvector.

\textbf{Datasets. }Two datasets were employed to assess cross-domain adaptability: the primary SunDB dataset, comprising technical manuals, logs, and specifications with embedded tables and code, and a secondary Electrical dataset of manuals and schematics with diagrams. Each dataset included 245 expert-curated questions with ground-truth answers, annotated with source locations (chapter, section, page) for precise validation.

\textbf{Metrics and Tools. }The evaluation focused on four core metrics, each with a success threshold of 0.7 (score range: 0-1), to assess retrieval and generation quality:
\begin{itemize}
    \item \textbf{Answer Relevancy (AR)}: Measures alignment between the query and the generated answer, computed as:
        \[
        \text{AR} = \frac{1}{N} \sum_{i=1}^{N} \cos(E_{g_i}, E_o) = \frac{1}{N} \sum_{i=1}^{N} \frac{E_{g_i} \cdot E_o}{\|E_{g_i}\| \|E_o\|},
        \]
        where \( N=3 \), \( E_{g_i} \) is the embedding of the \( i \)-th generated answer, and \( E_o \) is the query embedding.
    \item \textbf{Contextual Recall (CR)}: Evaluates the retrieval of all relevant information:
        \[
        \text{CR} = \frac{\text{Number of Attributable Statements}}{\text{Total Number of Statements}}.
        \]
    \item \textbf{Contextual Precision (CP)}: Assesses the accuracy of retrieved context, excluding irrelevant data:
        \[
        \text{CP@K} = \frac{\sum_{k=1}^K (\text{Precision@k} \times v_k)}{\text{Total number of relevant items in top } K \text{ results}},
        \]
        where \( \text{Precision@k} = \frac{\text{true positives@k}}{\text{true positives@k} + \text{false positives@k}} \), and \( v_k \in \{0,1\} \) indicates relevance.
    \item \textbf{Faithfulness (F)}: Measures how accurately the answer reflects the retrieved context:
        \[
        \text{F} = \frac{\text{Number of Truthful Claims}}{\text{Total Number of Claims}}.
        \]
\end{itemize}
These metrics were computed using RAGAS~\cite{ragas} for granular insights, DeepEval~\cite{langfuse} for end-to-end analysis, and LangFuse~\cite{langfuse} for detailed trace analysis, logging operations from query decomposition to answer generation.

\textbf{Baseline. }Two baseline comparisons were established. Externally, DO-RAG, implemented as SunDB.AI, was compared against FastGPT, a versatile open-source RAG framework with LLM-driven knowledge base construction; TiDB.AI, a graph-enhanced RAG framework tailored to database domain knowledge; and Dify.AI, an industrial-grade open-source RAG platform with generalized knowledge graph integration. Internally, DO-RAG with knowledge graph integration was contrasted against a variant relying solely on vector-based retrieval, isolating the knowledge graph’s impact.

\textbf{Evaluation Methodology. }The evaluation followed a structured methodology. Identical knowledge bases were constructed for all frameworks using the SunDB and Electrical datasets. The same 245 questions were tested across frameworks and domains. Three language models---DeepSeek-R1, DeepSeek-V3~\cite{deepseekai2024deepseekv3technicalreport}, and GPT-4o-mini~\cite{gpt4omini}---were selected. Metrics were computed for each model, and a composite score, averaging the four core metrics, provided a balanced performance measure.

\subsection{Results}

\textbf{External Baseline Comparison. }Table~\ref{tab:framework_comparison} presents the composite scores for the external baseline comparison, evaluating SunDB.AI against FastGPT, TiDB.AI, and Dify.AI across the tested language models. SunDB.AI consistently outperformed all three baselines.

\begin{table*}[h]
    \centering
    \caption{Comparative performance of graph-enhanced RAG frameworks (External Baseline).}
    \label{tab:framework_comparison}
    \resizebox{\textwidth}{!}{%
    \begin{tabular}{lccccccc}
        \toprule
        \multirow{2}{*}{\textbf{Model}} & \multicolumn{4}{c}{\textbf{Composite Score}} & \multicolumn{3}{c}{\textbf{Improvement over SunDB.AI}} \\
        \cmidrule(lr){2-5} \cmidrule(lr){6-8}
        & \textbf{SunDB.AI} & \textbf{FastGPT} & \textbf{TiDB.AI} & \textbf{Dify.AI} 
        & \textbf{vs FastGPT} & \textbf{vs TiDB.AI} & \textbf{vs Dify.AI} \\
        \midrule
        GPT-4o-mini & \underline{\textbf{0.741}} & 0.728 & 0.597 & 0.629 
                   & $\uparrow$ 1.70\% & $\uparrow$ 24.02\% & $\uparrow$ 17.72\% \\
        DeepSeek-V3 & \underline{\textbf{0.704}} & 0.623 & 0.704 & 0.601 
                   & $\uparrow$ 13.10\% & $\uparrow$ 0.10\% & $\uparrow$ 17.31\% \\
        DeepSeek-R1 & \underline{\textbf{0.847}} & 0.659 & 0.693 & 0.635 
                   & $\uparrow$ 28.50\% & $\uparrow$ 22.21\% & $\uparrow$ 33.38\% \\
        \bottomrule
    \end{tabular}%
    }
\end{table*}

\begin{figure}[h]
    \centering
    \includegraphics[width=0.99\linewidth]{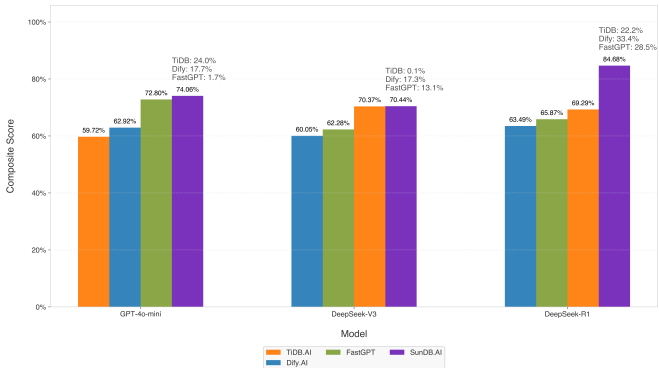}
    \caption{Composite score comparison across frameworks and language models.}
    \label{fig:framework_comparison_chart}
\end{figure}

Figure~\ref{fig:framework_comparison_chart} visualizes these results, illustrating SunDB.AI’s consistent superiority.

\textbf{Internal Baseline Comparison. }Table~\ref{tab:kg_impact} evaluates the impact of knowledge graph integration using DeepSeek-R1 and DeepSeek-V3. With the knowledge graph, both models achieved perfect Contextual Recall (1.000), with DeepSeek-V3 improving Answer Relevancy by 5.7\% and Contextual Precision by 2.6\%. Without the knowledge graph, recall dropped to 0.964–0.977, and Faithfulness was lower due to reliance on unstructured vector search. DeepSeek-R1 exhibited a slight Faithfulness decline (-5.6\%) with the knowledge graph, likely due to creative deviations.

\begin{table*}[h]
    \centering
    \caption{Impact of knowledge graph integration on DeepSeek models.}
    \label{tab:kg_impact}
    \resizebox{\textwidth}{!}{%
    \begin{tabular}{lcccccc}
        \toprule
        \textbf{Metric} & \multicolumn{2}{c}{\textbf{DeepSeek-R1}} & \multicolumn{2}{c}{\textbf{DeepSeek-V3}} & \textbf{KG Impact (R1)} & \textbf{KG Impact (V3)} \\
        & \textbf{Baseline} & \textbf{+KG} & \textbf{Baseline} & \textbf{+KG} & \textbf{($\Delta$)} & \textbf{($\Delta$)} \\
        \midrule
        Answer Relevancy & 0.851 & 0.820 & 0.871 & \underline{\textbf{0.921}} & $\downarrow$ 3.6\% & $\uparrow$ 5.7\% \\
        Contextual Recall & 0.964 & \underline{\textbf{1.000}} & 0.977 & \underline{\textbf{1.000}} & $\uparrow$ 3.7\% & $\uparrow$ 2.4\% \\
        Contextual Precision & 0.907 & \underline{\textbf{0.918}} & 0.919 & \underline{\textbf{0.943}} & $\uparrow$ 1.2\% & $\uparrow$ 2.6\% \\
        Faithfulness & 0.718 & 0.678 & 0.711 & \underline{\textbf{0.714}} & $\downarrow$ 5.6\% & $\uparrow$ 0.4\% \\
        \bottomrule
    \end{tabular}%
    }
\end{table*}

\textbf{Domain-Specific Performance. }Tables~\ref{tab:performance_database} and~\ref{tab:performance_electrical} display the performance metrics for each language model in the SunDB and Electrical domains, respectively. In both domains, Contextual Recall values are at or near 1.0. Variations in Answer Relevancy, Contextual Precision, and Faithfulness reveal model-specific strengths.

\begin{table*}[h]
    \centering
    \caption{Performance comparison of LLMs in the database domain.}
    \label{tab:performance_database}
    \resizebox{0.99\textwidth}{!}{
    \begin{tabular}{lcccc}
        \toprule
        \textbf{Model} & \textbf{Answer Relevancy} & \textbf{Contextual Recall} & \textbf{Contextual Precision (ragas)} & \textbf{Faithfulness} \\
        \midrule
        DeepSeek-R1   & 0.820407 & \underline{\textbf{1.000000}} & 0.918463 & 0.677958 \\
        DeepSeek-V3   & 0.920967 & \underline{\textbf{1.000000}} & 0.942694 & 0.714305 \\
        GPT-4o        & \underline{\textbf{0.944203}} & 0.979167 & 0.950279 & 0.770946 \\
        GPT-4o mini   & 0.939535 & 0.977273 & 0.921744 & \underline{\textbf{0.842160}} \\
        Grok 3        & 0.898365 & \underline{\textbf{1.000000}} & \underline{\textbf{0.953021}} & 0.836704 \\
        \bottomrule
    \end{tabular}
    }
\end{table*}

\begin{table*}[h]
    \centering
    \caption{Performance comparison of LLMs in the electrical domain.}
    \label{tab:performance_electrical}
    \resizebox{0.99\textwidth}{!}{
    \begin{tabular}{lcccc}
        \toprule
        \textbf{Model} & \textbf{Answer Relevancy} & \textbf{Contextual Recall} & \textbf{Contextual Precision (ragas)} & \textbf{Faithfulness} \\
        \midrule
        DeepSeek-V3  & 0.965342 & \underline{\textbf{1.000000}} & 0.919427 & 0.912089 \\
        GPT-4o       & \underline{\textbf{0.975000}} & \underline{\textbf{1.000000}} & \underline{\textbf{0.930203}} & 0.870054 \\
        GPT-4o mini  & 0.958386 & \underline{\textbf{1.000000}} & 0.913326 & \underline{\textbf{0.943798}} \\
        O3 mini      & 0.949360 & \underline{\textbf{1.000000}} & 0.888545 & 0.880241 \\
        \bottomrule
    \end{tabular}
    }
\end{table*}

\subsection{Discussion}

\textbf{Limitations. }The framework’s reliance on language models presents challenges, as creative models like DeepSeek-R1 occasionally introduced hallucinations despite knowledge graph grounding. The dataset, limited to 245 questions per domain, may not capture rare or edge-case queries, potentially limiting generalizability. The computational overhead of multi-agent extraction and hybrid retrieval, while optimized, remains significant for real-time updates in large-scale deployments.

\textbf{Future Work. }Future efforts will focus on enhancing hallucination mitigation through stricter prompt engineering to prioritize factual adherence. Expanding datasets to include diverse, edge-case queries will improve robustness. Distributed processing and adaptive caching mechanisms will be explored to enhance scalability and reduce latency.

\section{Conclusion}\label{sec5:conclusion}
This paper introduces DO-RAG, a domain-specific retrieval-augmented generation framework that addresses limitations of existing RAG systems in closed-domain QA. DO-RAG transforms unstructured, multimodal domain data into dynamic, multi-level knowledge graphs using an agentic chain-of-thought extraction pipeline. It integrates graph traversal with semantic vector search to retrieve context-rich, structurally grounded information. A post-generation refinement step cross-verifies outputs against the knowledge graph, iteratively correcting hallucinations to enhance factual accuracy. Empirical results in the database and electrical domains demonstrate near-perfect recall and answer relevancy exceeding 94\%, with DO-RAG outperforming baseline frameworks by up to 33.38\%. These findings demonstrate DO-RAG’s effectiveness in delivering robust, high-accuracy QA across specialized domains, unifying structured knowledge representation with generative reasoning for scalable and adaptable information systems.



\printbibliography

\end{document}